# DISentangled Counterfactual Visual interpretER (DISCOVER) generalizes to natural images

Oded Rotem[1] and Assaf Zaritsky[1]

[1]Department of Software and Information Systems Engineering, Ben-Gurion University of the Negev, Beer-Sheva 84105, Israel

Assaf Zaritsky, assafzar@gmail.com

## Abstract

We recently presented *DISentangled COunterfactual Visual interpretER* (*DISCOVER)*, a method toward systematic visual interpretability of image-based classification models and demonstrated its applicability to two biomedical domains. Here we demonstrate that DISCOVER can be applied to the domain of natural images. First, DISCOVER visually interpreted the nose size, the muzzle area, and the face size as semantic discriminative visual traits discriminating between facial images of dogs versus cats. Second, DISCOVER visually interpreted the cheeks and jawline, eyebrows and hair, and the eyes, as discriminative facial characteristics. These successful visual interpretations across two natural images domains indicate that DISCOVER is a generalized interpretability method.

# Introduction

Understanding the decision process of machine learning predictions with explainable artificial intelligence (XAI) is key for providing transparency, avoiding spurious biases in the model's decision process, and suggesting clues in cases of erroneous predictions (Ribeiro et al, 2016, Mengnan et al, 2019). XAI is especially critical in biomedical imaging domains that are hard to interpret and where novel interpretations can lead to generation of specific hypotheses regarding the underlying biological mechanisms driving the model's prediction We recently reported a new counterfactual-based method for visual interpretability of image-based classification models called *DISentangled COunterfactual Visual interpretER* (*DISCOVER*) (Rotem et al., bioRxiv). The main advantage of DISCOVER is learning a disentangled latent image representation constraining each latent feature to represent a distinct property in the image space. This property of the latent representation, known as "factor of variation" (Mathieu et al, 2016), enables visual counterfactual-based traversal and interpretation of the latent space. We previously demonstrated that DISCOVER can be used to visually interpret two biomedical imaging classification models: distinguishing between high and low in vitro fertilization (IVF) embryo morphological quality and distinguishing between healthy and Alzheimer's-diseased brains in MRI images (Rotem et al., bioRxiv). Here, we turned to the domain of natural images to showcase the generalization of DISCOVER toward visual interpretability of image-based classification models beyond the biomedical domain. First, DISCOVER visually interpreted the nose size, the muzzle area, and the face size as semantic discriminative visual traits discriminating between facial images of dogs versus cats. Second, DISCOVER visually interpreted the cheeks and jawline, eyebrows and hair, and the eyes as semantic discriminative visual traits discriminating between human male and female facial images. We conclude that DISCOVER is a generalized visual interpretability method.

# Methods

**DISCOVER architecture and optimization**

DISCOVER's architecture and optimization is thoroughly described in (Rotem et al., bioRxiv). Here we briefly describe the model's high-level architecture and optimization. The DISCOVER architecture is composed of 3 modules and is trained to simultaneously optimize six loss terms. The first module (loss terms #1-2) uses adversarial autoencoding ([Makhzani et al. 2015](#)) and perceptual loss minimization ([Pihlgren et al. 2020)](#) for high quality reconstruction and smooth and realistic traversal of the latent space through its reconstructed images. The second module (loss term #3) enforces a domain-specific classification-oriented encoding by minimizing the discrepancy between the binary classifier scores of the input image and its corresponding reconstructed image. The third module (loss terms #4-6) enforces a classification-driving counterfactual disentangled representation by decorrelating the latent features and by associating a subset of the latent features with the classifier that is being interpreted. After DISCOVER was trained, it can be visually interpreted by following the "*visual counterfactual alteration*", a Structural Similarity Index (SSIM) ([Renieblas et al. 2017)](#)-based visualization.

**Code and data availability**

DISCOVER source code (Python with Tensorflow 2.2) is publicly available, [https://github.com/zaritskylab/DISCOVER](https://github.com/zaritskylab/DISCOVER), and includes a trained model for gender classification and its corresponding DISCOVER model. The celebA dataset is available [https://mmlab.ie.cuhk.edu.hk/projects/CelebA.html](https://mmlab.ie.cuhk.edu.hk/projects/CelebA.html).

# Results

**DISentangled COunterfactual Visual interpretER - DISCOVER**

We previously reported *DISCOVER*, a method designed to interpret the visual properties underlying a binary classifier's decision. Briefly explained, DISCOVER is a generative model that is trained to transform an input image to a latent space. This latent image representation is optimized toward effective visual interpretation of the classifiers decision by constraining the representation toward the following properties: (1) high-quality and realistic reconstruction and latent space traversal, (2) domain-specific classification oriented encoding (i.e., similar classification of observed and reconstructed images), and (3) a subset of latent features optimized toward explainability by visual disentanglement and correlation with the classifier that is being interpreted. These properties, and especially the latent space disentanglement, enable enhanced interpretability by iterative traversal over the latent space, one latent feature at a time. The corresponding counterfactual explanations generate images where specific classification-driving image properties are exaggerated, while maintaining the other image properties unchanged (Fig. 1). This traversal over disentangled latent features that amplify one specific image property at a time in the image space enable easier human interpretation. Full details can be found in (Rotem et al., bioRxiv).

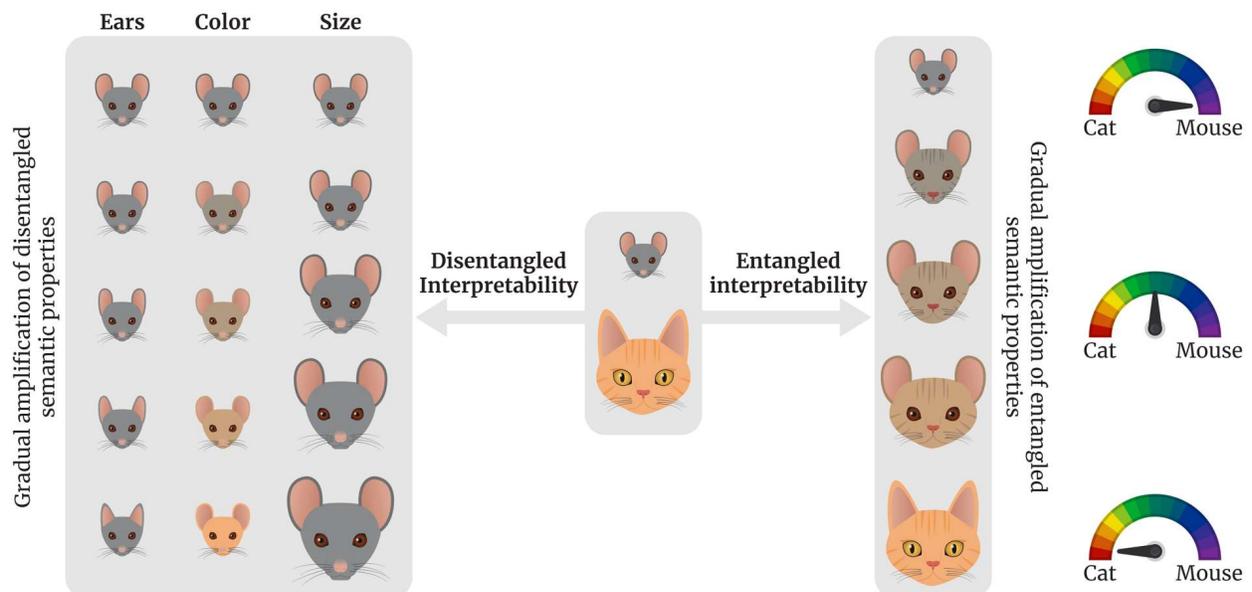

**Figure 1. Entangled vs disentangled interpretability representation**. A cartoon depicting counterfactual explanations using disentangled (left) versus entangled (right) latent representations.

Interpreting a mouse/cat classifier via counterfactual explanation by gradually morphing an image from mouse to cat. The red-to-blue gauge on the right indicates the classifier's prediction. Left: disentangled representation enables to amplify one classification-driving semantic image property at a time. Right: entangled representation causes simultaneous morphing of multiple classification-driving semantic image properties, hampering interpretation.

**Visual interpretation of classification-driving features distinguishing between dog and cat images**

The AFHQ dataset ([Choi et al. 2020](#)) contains 15K aligned and cropped RGB images (512x512 pixels) with an associated cat/dog/wildlife attribute. We focused on the binary classification of dog versus cat images with 5,000 images per class. We trimmed 20 pixels from each side to remove background nuisance and the image was then resized to 64x64 pixels. All images were converted to grayscale and divided by 255 to the range [0-1]. A VGG-19 classification model was fine-tuned to discriminate between dog and cat face images, we call this model ANIMAL-CLF. The training followed the same procedure described in (Rotem et al., bioRxiv) except the inclusion of left-right flipping augmentations. The AUC for the test data (200 images for each class) was 0.95 (Fig. 2A).

We trained DISCOVER using the trained classifier ANIMAL-CLF and the same training dataset and interpreted the top three ranked latent features according to their association with the classifier's score. Namely, latent features #6, #1, and #3, with Pearson correlation coefficient of 0.54, 0.53, and 0.50, respectively (Fig. 2B). Visualization of the counterfactual alteration revealed that feature #6 encoded the nose size (larger for dog), feature #1 the muzzle area (only for dogs), and feature #3 the face size (longer for dogs) (Fig. 2C).

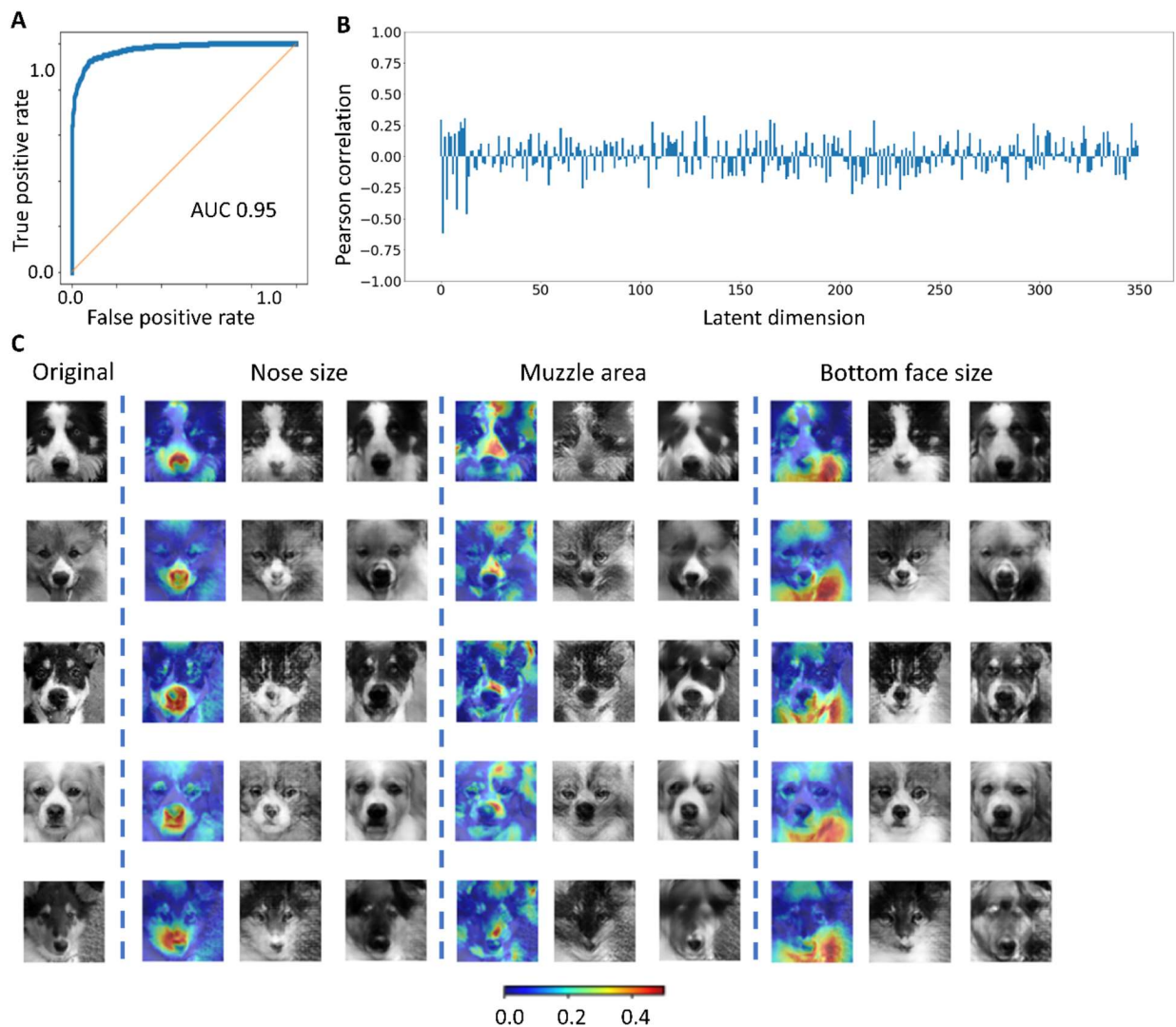

Figure 2. interpreting dog-versus-cat classification. Data from the public AFHQ dataset (Choi et al. 2020). (**A**) ROC curve of the ANIMAL-CLF with train/test sets of 4800/200 for every class, test AUC = 0.95. (**B**) Pearson correlation coefficient (y-axis) between each latent feature (x-axis) and the ANIMAL-CLF's classification score. (**C**) Five examples showing the three main facial features which alter the animal type. The first column presents the original image. Second, third and fourth main columns (separated by dashed lines) correspond to latent features #6, #1 and #3 with Pearson correlation coefficient of 0.64, 0.53, and 0.49, respectively. Each column shows the visual counterfactual alteration (left), cat-altered image (middle) and dog-altered image (right) when traversing the latent feature by +3 or -3 std (depending on the correlation sign). Posting these photos does not conflict with the copyright policy of the database

**Visual interpretation of classification-driving features distinguishing between male and female facial images**

The celebA dataset ([Liu et al. 2015](#)) contains 202,599 aligned and cropped RGB images (64x64 pixels) of 10,000 celebrities' faces with an associated male/female attribute (as well as additional 40 binary annotations such as smile, hat etc.). We trimmed 15 pixels from each side to remove background nuisance and the image was then resized back to 64x64 pixels. All images were converted to grayscale and divided by 255 to the range [0-1]. A VGG-19 classification model was fine-tuned to discriminate between male and female face images, we call this model GENDER-CLF. The training followed the same procedure described in (Rotem et al., bioRxiv) with no augmentations. 15,000 images from each gender were randomly selected for training. The AUC for the test data (1,000 images for each gender) was 0.96 (Fig. 3A).

We trained DISCOVER using the trained face classifier GENDER-CLF and training dataset composed of 164,268 (65,183 male and 99,085 female) images without augmentations and interpreted the top three ranked latent features, namely #2, #4, and #3, with Pearson correlation coefficient of 0.68, 0.49, and 0.42, respectively (Fig. 3B). Visualization of the counterfactual alteration revealed that feature #2 encoded the cheeks and jawline (smaller face for females), feature #4 the eyebrows and hair (thinner hair for females), and feature #3 the eyes (darker for females) (Fig. 3C). These traits could not be attained with GradCAM (Fig. 3C, right) and were consistent with previous studies that highlighted cheeks, eyes and eyebrows as discriminative facial characteristics ([Bannister et al. 2022](#), Xia et al. 2018).

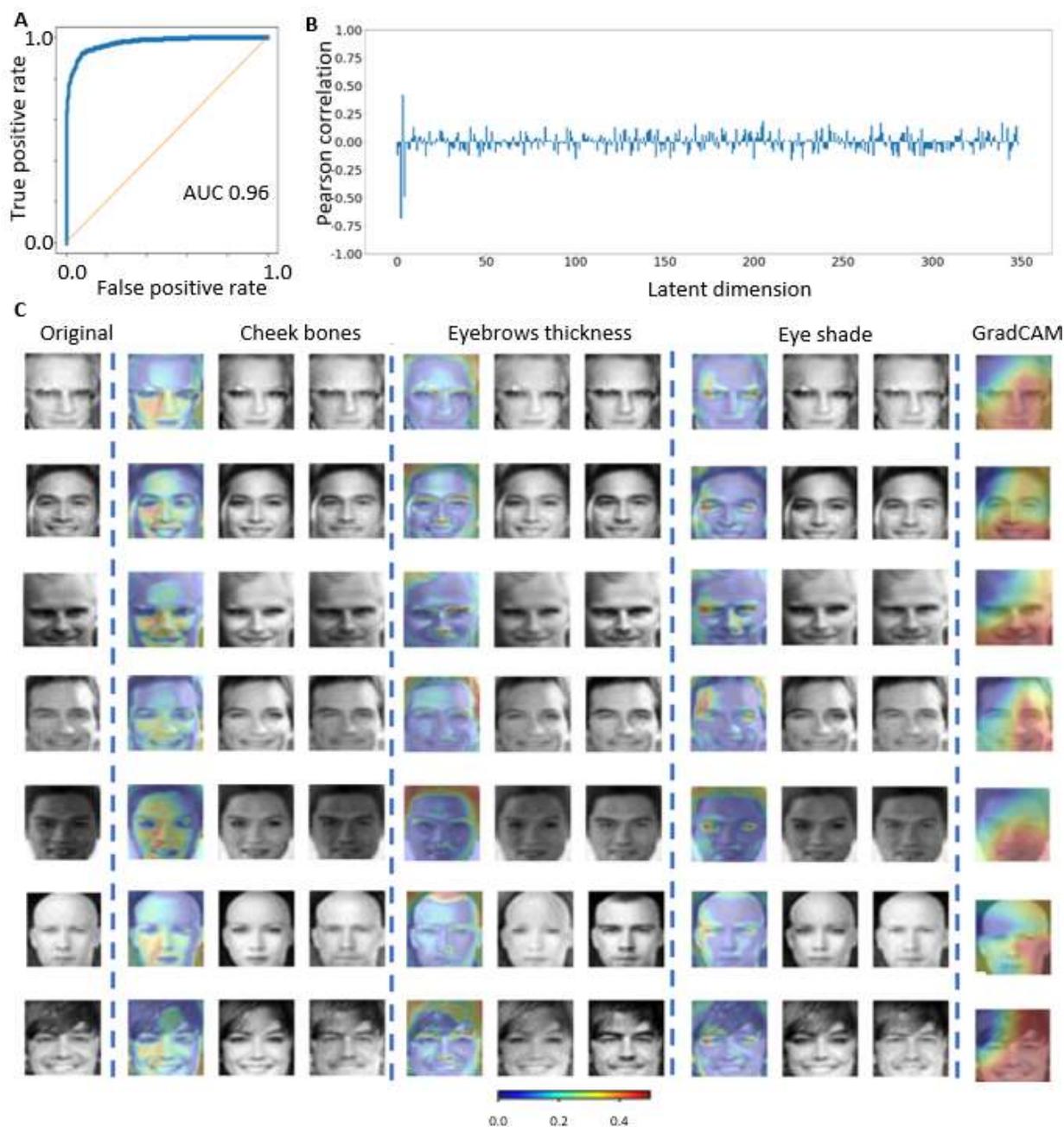

**Figure 3. Interpreting gender classification.** Data from the public CelebA dataset, https://mmlab.ie.cuhk.edu.hk/projects/CelebA.html. (**A**) ROC curve of the GENDER-CLF with train/test sets of 15K/1K for female and for male images, test AUC 0.96. (**B**) Pearson correlation coefficient (y-axis) between each latent feature (x-axis) and the GENDER-CLF's classification score, using a test set of 2K faces that were not used to train DISCOVER (trained with 164,268 images). (**C**) seven examples showing the three main facial features which alter the gender of the face. First column presents the original image. Second, third and fourth main columns (separated by dashed lines) correspond to latent features #2, #4 and #3 with Pearson correlation coefficient of 0.68, 0.49, and 0.42, respectively. Each column shows the visual counterfactual alteration (left), female-altered image (middle) and male-altered image (right) when traversing the latent feature by +3 or -3 std (depending on the correlation sign). The last column shows the Gradcam heatmap for each example. Posting these photos does not conflict with the copyright policy of the database.

## Summary

Demonstrating applicability to two biomedical (IVF, Alzheimer's) (Rotem et al., bioRxiv) and two general computer vision (gender, animal) datasets indicate that DISCOVER is a generalized interpretability method.

## Funding and Acknowledgements

This research was supported by the Israel Council for Higher Education (CHE) via the Data Science Research Center, Ben-Gurion University of the Negev, Israel (to AZ), and by the Rosetrees Trust (to AZ).

## Author Contribution

OR and AZ conceived the study. OR analyzed the data. OR and AZ interpreted the data, drafted the manuscript. AZ mentored OR.

## Competing Financial Interests

The authors declare no financial interests.